# Amharic-English Cross-Language Question Answering Using Statistical Machine Translation, Named Entity Recognition, and Semantic Indexing

Emebet Bekele and Dr. Fekade Getahun

**Abstract**

The predominance of English-language Web content creates an information-access barrier for speakers of low-resource languages such as Amharic. This paper presents the design, implementation, and evaluation of a bidirectional Amharic-English cross-language question answering (CLQA) system that allows a user to formulate a natural-language question in either Amharic or English and retrieve an answer from documents written in the other language. The system integrates a statistical machine translation (SMT) component for query translation, a maximum-entropy named entity recognition (NER) component for Amharic text processing, semantic-based indexing, question analysis, passage retrieval, and answer extraction. A domain-diverse Amharic-English parallel corpus was used to train the SMT component, while monolingual corpora supported language modeling. Documents were normalized, tokenized, stemmed, indexed, and semantically enriched before retrieval. The original evaluation used precision and recall as the principal effectiveness measures. For Amharic-to-English retrieval, the system achieved 72% precision and 79% recall; for English-to-Amharic retrieval, it achieved 65% precision and 70% recall. These findings show that a modular translation-and-retrieval architecture can provide useful cross-language access even under low-resource conditions. This revised article preserves the original experimental claims while situating them in the context of subsequent multilingual transformer and Amharic QA research.

**Author affiliation:** Addis Ababa University, Addis Ababa, Ethiopia

**Provenance note.** An earlier thesis version of this work, *Design and Development of Amharic-English Cross Language Question Answering System*, was deposited at Addis Ababa University in 2019 [1]. This article restructures that work as a research paper and adds post-2019 literature context. It does **not** claim new experiments beyond the original study.

## 1. Introduction

Question answering (QA) aims to satisfy an information need by returning a concise answer to a natural-language question rather than merely presenting a ranked list of documents. In conventional information retrieval (IR), a user must inspect retrieved documents and determine which passages contain the needed information. QA systems add linguistic analysis and answer extraction so that the final output is closer to the user's actual information need [2].

This distinction becomes particularly important in multilingual and low-resource environments. A large share of information available on the Web is written in English, whereas many users formulate information needs most naturally in another language. For Amharic speakers, this mismatch creates two related barriers: the user may not be comfortable composing an English query, and relevant information may exist only in English even when an Amharic information need is clear. Cross-language information retrieval (CLIR) and cross-language question answering (CLQA) address this problem by bridging the language of the user's question and the language of the document collection.

Amharic is a Semitic language written primarily in the Ethiopic script. It is morphologically rich, has spelling and orthographic variation, and historically has had substantially fewer computational linguistic resources than English. Earlier Amharic IR and QA research therefore relied heavily on normalization, stemming, dictionaries, rules, statistical learning, and task-specific linguistic resources [3-5]. These characteristics motivated the architecture developed in this research.

The system presented here is bidirectional: it supports an Amharic question against English documents and an English question against Amharic documents. Rather than translating an entire document collection, the architecture follows a query-translation strategy. A user's question is analyzed, translated into the target document language, and submitted to an indexed document collection. Candidate passages are then ranked, named entities and expected answer types are used to narrow the search space, and a final answer is selected.

The original study was carried out in the statistical machine translation era and used the Moses SMT toolkit [6]. It also introduced an Amharic NER component based on maximum-entropy classification [7] and used semantic-based indexing to improve retrieval specificity. The evaluation reported 72% precision and 79% recall for Amharic-to-English retrieval, and 65% precision and 70% recall for English-to-Amharic retrieval [1].

The research makes the following contributions:

1. It defines a modular architecture for bidirectional Amharic-English CLQA.
2. It applies statistical query translation to bridge Amharic and English document collections.
3. It integrates an Amharic maximum-entropy NER component into retrieval and answer extraction.
4. It combines lexical indexing with semantic categories and expected answer types to reduce irrelevant candidates.
5. It evaluates both translation directions using precision and recall and documents direction-specific error patterns.
6. In this revised presentation, it positions the original pipeline relative to later multilingual language models and newer Amharic QA resources without altering the original experimental results.

The remainder of the paper is organized as follows. Section 2 reviews related work. Section 3 defines the research problem and system objectives. Section 4 presents the architecture. Section 5 details the methodology. Section 6 describes the evaluation and results. Section 7 discusses the findings and their present-day relevance. Section 8 summarizes limitations and threats to validity, Section 9 identifies future directions, and Section 10 concludes.

# 2. Background and Related Work

## 2.1 Question answering and information retrieval

IR systems rank documents or passages in response to a query. Classical IR pipelines typically include document acquisition, preprocessing, indexing, query processing, matching, ranking, and evaluation [2]. QA extends this pipeline with question classification, expected answer type detection, candidate answer extraction, and answer ranking. The distinction is operationally important: a retrieval engine may return a highly relevant document while a QA system must still identify the specific phrase, entity, date, quantity, or passage that satisfies the question.

Cross-language retrieval introduces an additional transformation between the user query and the target collection. Translation can be performed at the query, document, or representation level. Query translation is attractive in low-resource settings because it avoids translating an entire document collection. However, errors in translating even one key query term can severely affect retrieval, especially for short questions containing names, rare terms, or morphologically complex words.

## 2.2 Amharic-English cross-language retrieval

Prior Amharic-English retrieval research demonstrated that cross-language access is feasible but sensitive to translation resources. Argaw [3], for example, evaluated Amharic-English CLIR using dictionary-based query translation, pseudo-relevance feedback, and alternative word-sense handling strategies. Such work established several practical issues that remain relevant to Amharic cross-language access: out-of-vocabulary terms, named entities, morphological variation, and ambiguity in dictionary or machine translation output.

The present work differs from document-oriented CLIR by targeting question answering. The system must not only retrieve a relevant document, but also analyze a natural-language question, infer the expected answer category, retrieve candidate passages, and identify the most likely answer.

## 2.3 Earlier Amharic question answering

Yimam and Libsie developed TETEYEQ, an early Amharic factoid QA system [4]. Their work emphasized document normalization, question classification, document retrieval, and answer selection, showing that language-specific preprocessing is central to Amharic QA. Taffa and Libsie later addressed Amharic biography, definition, and description questions using a combination of rule-based and machine-learning techniques [5]. These systems were primarily monolingual: the question and target documents were in Amharic.

The present research extends the problem in a different direction by introducing a language boundary between the question and the answer source. This adds translation quality as a first-order determinant of downstream QA performance.

## 2.4 Statistical machine translation

Statistical machine translation models translation as a probabilistic decision problem. Phrase-based SMT became a widely used practical approach, with Moses providing an open-source toolkit for training and decoding phrase-based models [6]. The underlying statistical view builds on earlier probabilistic translation modeling, including the IBM-model tradition [14].

For CLQA, SMT is particularly useful as a query translation component. The translation model can exploit parallel sentence pairs rather than relying only on bilingual dictionary entries. Nevertheless, SMT performance in a low-resource language pair remains strongly dependent on parallel-corpus coverage. Short questions are especially vulnerable to mistranslation because they provide less contextual evidence for disambiguation.

### 2.5 Named entity recognition and semantic indexing

Many factoid questions target named entities such as people, places, organizations, dates, or quantities. NER therefore provides a natural bridge between question analysis and answer extraction. The original system developed an Amharic named entity tagger using a maximum-entropy classifier, a well-established log-linear modeling framework for NLP [7]. The detected entities and semantic categories were then used during indexing and answer selection.

Semantic-based indexing in this work does not replace lexical indexing. Instead, it enriches the index with information that can help reject lexically matched but semantically incompatible candidates. For example, if a question expects a person, passages containing recognized person entities can be favored over passages whose relevant terms occur only in unrelated contexts.

### 2.6 Post-2019 multilingual and Amharic NLP context

The technological landscape changed substantially after the original study. Multilingual representation learning made it possible to transfer linguistic knowledge across many languages. XLM-R demonstrated strong cross-lingual transfer across 100 languages and improved performance on multilingual NER and question answering benchmarks [9]. mT5 extended the text-to-text paradigm to 101 languages [10]. For African languages, AfriBERTa showed that multilingual models trained specifically on low-resource African languages can be competitive despite relatively small training corpora [11].

Question answering resources also improved. TyDi QA established a multilingual information-seeking QA benchmark covering typologically diverse languages [8]. For Amharic specifically, Taffa, Usbeck, and Assabie later released a publicly available QA benchmark containing 2,628 question-answer pairs drawn from 378 Amharic Wikipedia articles and evaluated an XLM-R-based reader [13]. Amharic NER resources also advanced: Jibril and Tantuğ introduced an annotated Amharic NER corpus and a neural recognizer [12].

These later developments should not be interpreted as components of the 2019 system. Rather, they show how the original modular CLQA problem can now be revisited with stronger multilingual encoders, neural machine translation, dense retrieval, and public benchmarking datasets. The original pipeline remains useful as a transparent classical baseline and as evidence of the value of explicit language-specific preprocessing under resource constraints.

## 3. Problem Definition and Research Objectives

Let the source language of a user question be denoted by $L_s$ and the language of the target document collection by $L_t$, where $L_s$, $L_t \in$ {Amharic, English} and $L_s \neq L_t$. Given a natural-language question $q$ in $L_s$, the system must identify an answer $a$ contained in or supported by documents written in $L_t$.

The task can be decomposed into five linked problems:

1. **Language-specific processing:** normalize and analyze the source question and target documents.
2. **Cross-language mapping:** translate the question into the target language while preserving content words, named entities, and question intent.
3. **Retrieval:** identify documents and passages likely to contain the answer.

4. **Semantic compatibility:** determine the expected answer type and identify candidate entities or expressions of that type.
5. **Ranking and extraction:** select the candidate that best satisfies the question.

The general objective of the research is therefore to design and evaluate a bidirectional Amharic-English CLQA architecture under low-resource conditions.

The specific objectives are to:

- develop an Amharic-English query translation model using SMT;
- preprocess and index Amharic and English documents using language-appropriate pipelines;
- develop an Amharic NER component using maximum-entropy classification;
- classify questions by expected answer type;
- combine keyword evidence and semantic categories during indexing and retrieval;
- retrieve candidate passages and extract ranked answers; and
- evaluate the effectiveness of both Amharic-to-English and English-to-Amharic retrieval directions.

# 4. System Architecture

The proposed system contains six principal components: (1) Web crawler, (2) indexer, (3) question translation module, (4) question analysis module, (5) passage retrieval module, and (6) answer extraction module [1]. Figure 1 summarizes the processing flow.

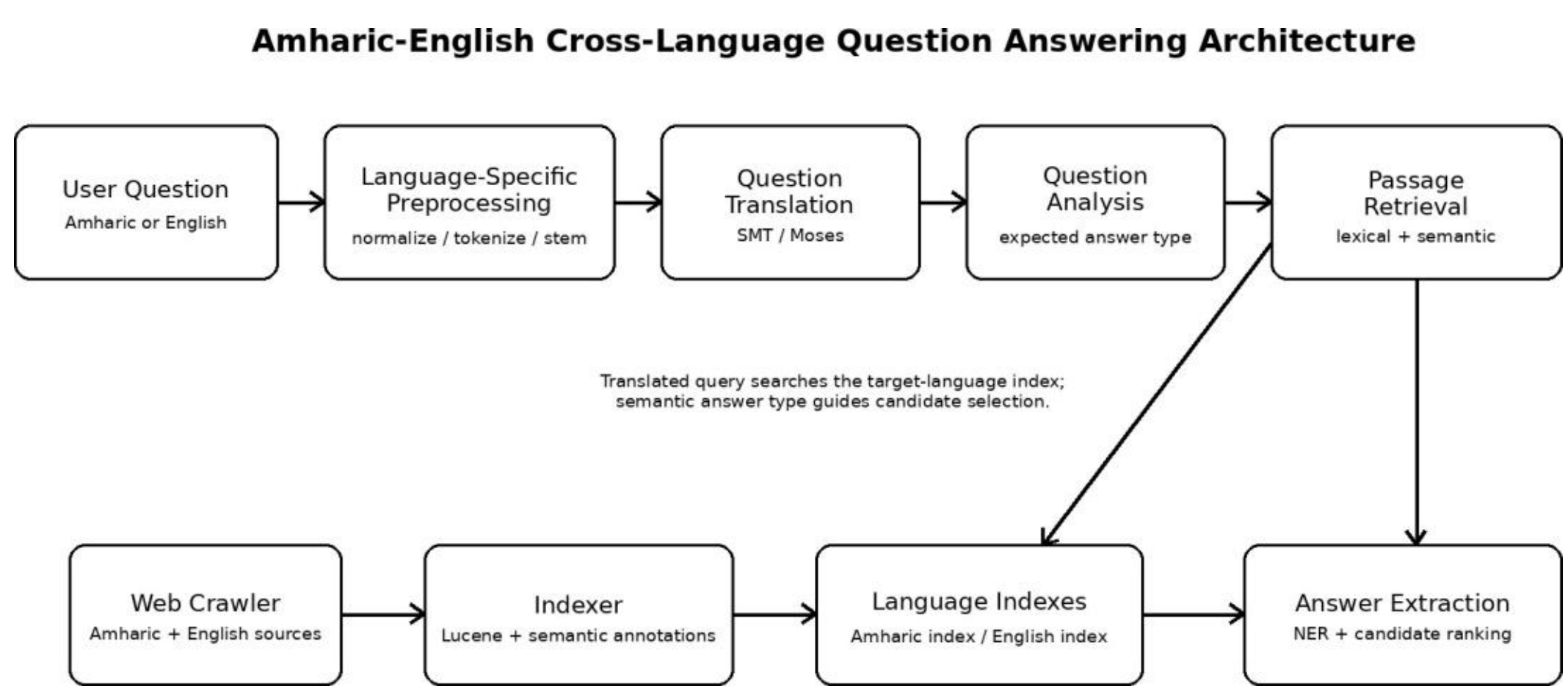

Figure 1: High-level architecture of the Amharic-English CLQA system.

**Table 1. Principal components and responsibilities.**

| Component | Main responsibility |
|---|---|
| Web crawler | Acquires Amharic and English documents from selected Web sources. |
| Indexer | Normalizes, tokenizes, removes stop words, stems, annotates named entities, and constructs searchable indexes. |
| Question translation | Translates the user's question into the target document language using the SMT model. |
| Question analysis | Parses the question, identifies important terms, and predicts the expected answer type. |
| Passage retrieval | Retrieves and ranks candidate passages using lexical and semantic evidence. |
| Answer extraction | Identifies candidate answers, checks semantic compatibility, and ranks the final answer. |

The architecture follows a query-translation model. The original question is retained for question analysis, while a translated form is produced for searching the target-language collection. This separation is important because some question-type cues may be clearer in the source language than after translation.

The system also maintains separate language-specific preprocessing pipelines. English and Amharic differ in tokenization behavior, morphology, orthographic normalization requirements, and resource availability. Treating preprocessing as language specific reduces the risk of applying assumptions from English directly to Amharic.

# 5. Methodology

## 5.1 Document collection and indexing

Documents were collected from Web sources and organized into Amharic and English collections. During indexing, each document passed through a language-specific pipeline consisting of tokenization, stop-word removal, normalization, stemming, and named entity recognition. Apache Lucene was used as the underlying indexing and retrieval technology in the original implementation [1].

The indexing strategy included both lexical and semantic information. Conventional lexical indexing preserves terms needed for keyword retrieval, while semantic annotations record entity categories and other information useful for matching an expected answer type.

## 5.2 Language preprocessing

Preprocessing is critical because surface-form variation can reduce both translation and retrieval performance. The Amharic pipeline normalizes orthographic variation before stemming and indexing. English text is tokenized, normalized, stripped of common stop words where appropriate, and stemmed.

The purpose of preprocessing is not merely to reduce index size. In a cross-language system, preprocessing affects translation-model lookup, query formulation, lexical overlap, and the downstream quality of candidate passages. Errors introduced early in the pipeline can therefore propagate to every later module.

### 5.3 Statistical machine translation

The translation component was developed with the Moses SMT toolkit [6]. Parallel Amharic-English sentence pairs were collected from multiple domains, including religious texts, legal documents, and news material [1]. Parallel data supported the translation model, while monolingual corpora were used for language modeling.

The training workflow included sentence preparation, tokenization, word or phrase alignment, translation-model estimation, language-model construction, and decoder configuration. At runtime, the source-language question is preprocessed and passed to the decoder. The output translation is then converted into a retrieval query for the target-language document index.

A query-translation architecture has a computational advantage over document translation because translation is performed only when a question is asked. Its main disadvantage is sensitivity to short-query ambiguity. A single mistranslated content word can substantially alter the retrieval results.

### 5.4 Named entity recognition

The original research developed an Amharic NER component using a maximum-entropy classifier [1,7]. The entity inventory included categories such as person, location, organization, and time. NER outputs were used both to enrich indexed documents and to support candidate answer extraction.

The maximum-entropy formulation is appropriate for combining heterogeneous contextual features in a probabilistic classifier. Although later Amharic NER systems use neural sequence models and transformer representations [12], the maximum-entropy component provided a practical supervised method under the computational and data constraints of the original study.

### 5.5 Question analysis and expected answer type

Question analysis transforms a natural-language question into a representation suitable for retrieval and answer extraction. The module performs question parsing, identifies informative terms, generates the retrieval query, and predicts an expected answer type.

A maximum-entropy classifier was used to classify questions into semantic categories such as person, location, time, and quantity [1]. This classification constrains answer extraction. For example, a question classified as seeking a location should preferentially rank candidate spans tagged as locations.

Expected-answer typing is especially valuable in CLQA because translation can increase lexical noise. Semantic compatibility provides an additional signal even when the translated query is imperfect.

### 5.6 Semantic-based indexing

The semantic indexing mechanism combines traditional term indexing with semantic categories identified during preprocessing. Rather than ranking passages solely by keyword overlap, the system can exploit the compatibility between the question's expected answer type and the entities contained in a passage.

Conceptually, a candidate passage $p$ receives evidence from at least two sources: lexical relevance to the translated query and semantic compatibility with the expected answer type. The original implementation used this combination operationally rather than as a single learned end-to-end scoring function. This modularity makes the contribution interpretable: improvements or errors can be traced to translation, retrieval, NER, question classification, or answer ranking.

### 5.7 Passage retrieval and answer extraction

The passage retrieval module searches the target-language index using the translated query and obtains a ranked set of candidate passages. The answer extraction module then inspects these candidates for entities or expressions compatible with the expected answer type.

Candidate answers are ranked according to retrieval relevance and semantic evidence. NER is particularly useful for factoid questions because it narrows the answer space to entities of the required class. If a high-ranking passage does not contain an entity of the expected category, another passage or candidate can be preferred.

This stage completes the CLQA pipeline: the final response is intended to be a concise answer rather than a list of documents.

## 6. Experimental Evaluation

### 6.1 Evaluation design

The original study evaluated the overall system in both cross-language directions: Amharic question to English documents and English question to Amharic documents [1]. Precision and recall were used as the principal effectiveness measures.

Precision is defined as the proportion of retrieved items judged relevant:

$$P = \frac{\text{relevant items retrieved}}{\text{items retrieved}}.$$

Recall is defined as the proportion of all relevant items that are retrieved:

$$R = \frac{\text{relevant items retrieved}}{\text{all relevant items}}.$$

For compact comparison in this revised article, an F1 value is also shown in Table 2. F1 was **not an additional original experiment**; it is mathematically derived from the reported precision and recall using $F1 = 2PR/(P+R)$.

### 6.2 Results

**Table 2. Reported cross-language retrieval results.**

| Retrieval direction | Precision | Recall | Derived F1* |
|---|---|---|---|
| Amharic question -> English documents | 72% | 79% | 75.3% |
| English question -> Amharic documents | 65% | 70% | 67.4% |

*Derived from the originally reported precision and recall; not separately reported in the 2019 study.

The results show that the system successfully retrieved relevant information in both directions, while the Amharic-to-English direction achieved higher precision and recall. The original analysis attributes part of this difference to the richer linguistic resources available for English and to translation and processing quality differences between the two directions [1].

### 6.3 Interpretation

The gap between directions is consistent with a pipeline in which translation and target-language processing are not equally resourced. In Amharic-to-English retrieval, the target collection benefits from comparatively mature English tokenization, normalization, stemming, and lexical resources. English-to-Amharic retrieval places more of the downstream burden on Amharic processing, where orthographic variation, morphological complexity, and limited annotated resources can create additional errors.

The results also indicate that translation quality alone is not sufficient. An imperfect translation may still produce a correct answer if the key content term is preserved and semantic indexing identifies a compatible candidate. Conversely, a fluent translation can fail if it mistranslates a named entity or changes a term that is crucial for retrieval.

## 7. Discussion

### 7.1 What the results demonstrate

The main finding is that a modular CLQA system can provide useful bidirectional access between a low-resource language and English without requiring the user to formulate the question in the document language. The achieved precision and recall values are meaningful because they reflect the combined behavior of translation, linguistic preprocessing, retrieval, NER, question analysis, and answer extraction rather than a single isolated classifier.

The architecture also demonstrates the value of explicit semantic constraints. A purely lexical cross-language retrieval system may return passages that contain translated query terms but do not contain an answer of the required type. Expected-answer classification and NER reduce this mismatch.

### 7.2 Principal error sources

The original study identifies three recurring error sources [1]:

- **translation ambiguity**, particularly when a source word can map to several target-language words;
- **named entity recognition errors**, which can prevent a correct candidate from receiving the expected semantic type; and
- **limited corpus coverage**, which restricts both translation-model learning and the range of retrievable evidence.

These errors interact. An incorrectly translated name may lead to retrieval failure before NER is ever applied. A correct translation may retrieve the right passage but still fail at answer extraction if the candidate is assigned the wrong entity type. Limited corpus coverage affects both model training and evaluation generality.

### 7.3 Relationship to modern multilingual QA

Modern multilingual transformer systems can replace several modules in the original architecture. Neural machine translation can replace phrase-based SMT; multilingual encoders can provide shared cross-language representations; dense retrievers can retrieve semantically related passages without exact lexical overlap; transformer readers can extract answer spans; and modern Amharic NER models can replace maximum-entropy tagging [9-13].

However, the classical architecture remains instructive for three reasons. First, each module is interpretable and can be evaluated independently. Second, the pipeline can operate with smaller task-specific datasets and modest computation. Third, its error decomposition remains useful when modern end-to-end systems fail: the same conceptual problems of translation, retrieval, entity typing, and answer selection still exist even when they are represented inside a large neural model.

A future neural replication of this study would therefore benefit from retaining the original bidirectional evaluation design while replacing individual components in controlled ablation experiments.

## 8. Limitations and Threats to Validity

Several limitations should be considered when interpreting the results.

First, the original evaluation predates widely adopted public Amharic QA benchmarks. Consequently, the reported scores are specific to the study's own data and evaluation setup and should not be compared directly with later benchmark scores such as those reported for Amh-QuAD [13].

Second, the available thesis-derived material used to prepare this revised article describes the training sources and methodology but does not expose all corpus-size counts, decoder hyperparameters, feature templates, or per-question evaluation records. This article therefore intentionally avoids inventing such values. A future reproducibility release should provide the original corpora, configuration files, evaluation questions, relevance judgments, and source code if they remain available.

Third, precision and recall provide a useful summary but do not capture every aspect of QA quality. Modern evaluations commonly include exact match, token-level F1, mean reciprocal rank, recall at $k$, latency, and statistical uncertainty. No claim is made here that the original evaluation satisfies present-day benchmarking standards.

Fourth, the document collections were drawn from particular Web and textual domains. Translation and QA performance may change on conversational text, social media, specialized technical domains, or text containing substantial spelling variation.

Fifth, the results reflect a pipeline architecture in which errors accumulate across modules. The study does not include a full ablation analysis that quantifies the independent contribution of SMT, NER, semantic indexing, and question classification.

These limitations do not invalidate the original result, but they define the scope within which it should be interpreted.

## 9. Future Work

The original research identified larger training datasets and stronger translation models as priorities. Subsequent developments make several concrete extensions possible.

A first direction is to replace phrase-based SMT with modern neural machine translation while preserving bidirectional evaluation. A second is to use multilingual encoders such as XLM-R for question representation, entity recognition, or answer extraction [9]. A third is to fine-tune sequence-to-sequence models such as mT5 for translation or generative QA [10].

A fourth direction is to train and evaluate on public Amharic QA resources such as the later Amharic benchmarking dataset [13]. This would make results directly reproducible and comparable. A fifth direction is to integrate newer Amharic NER resources such as ANEC [12] and assess whether improved entity recognition translates into better answer accuracy.

A sixth direction is hybrid retrieval. Sparse lexical retrieval can be combined with dense multilingual retrieval so that rare names and exact terms remain searchable while semantically related passages are also retrieved. Cross-encoder reranking could then prioritize the most relevant passages before answer extraction.

Finally, a modern replication should report module-level ablations, confidence intervals, per-question error categories, and public code/data artifacts. Such a replication would provide a direct bridge between the transparent 2019 pipeline and current multilingual neural QA.

## 10. Conclusion

This paper presented a bidirectional Amharic-English cross-language question answering architecture that combines statistical machine translation, language-specific preprocessing, maximum-entropy named entity recognition, semantic-based indexing, passage retrieval, and answer extraction. The central design principle is query translation: a user formulates a question in one language while the system searches documents in the other language and returns a concise answer.

The original evaluation reported 72% precision and 79% recall for Amharic-to-English retrieval and 65% precision and 70% recall for English-to-Amharic retrieval. The higher performance in the Amharic-to-English direction is consistent with differences in linguistic-resource availability and downstream processing quality. Error analysis identified translation ambiguity, NER errors, and limited corpus coverage as major constraints.

Although multilingual transformers and public Amharic QA datasets now provide stronger modeling options, the original architecture remains a useful low-resource baseline and a clear decomposition of the CLQA problem. Its modular structure makes the sources of error visible and provides a foundation for controlled comparison with neural translation, multilingual retrieval, and transformer-based readers.